\documentclass[]{jingdong}
\usepackage[toc,page,header]{appendix}

\usepackage{latexsym}
\usepackage[T1]{fontenc}
\usepackage[utf8]{inputenc}
\usepackage{microtype}
\usepackage{inconsolata}
\usepackage{graphicx}
\usepackage{hyperref}       
\usepackage{url}            
\usepackage{booktabs}       
\usepackage{amsfonts}       
\usepackage{nicefrac}       
\usepackage{stackengine}
\usepackage{microtype}      
\usepackage{colortbl}
\usepackage{xcolor}
\usepackage{amsmath}
\usepackage{amssymb}
\usepackage{amsthm}
\usepackage{mathrsfs}
\usepackage{pifont}
\usepackage{MnSymbol}
\usepackage{balance}
\usepackage{enumitem}
\usepackage{listings}
\usepackage{xcolor}
\usepackage{natbib}
\usepackage{multicol}

\AtBeginDocument{%
  \providecommand\BibTeX{{%
    \normalfont B\kern-0.5em{\scshape i\kern-0.25em b}\kern-0.8em\TeX}}}

\makeatletter
\DeclareRobustCommand\onedot{\futurelet\@let@token\@onedot}
\def\@onedot{\ifx\@let@token.\else.\null\fi}

\usepackage{setspace}
\usepackage{mathtools}

\usepackage{multirow,booktabs}
\usepackage{subcaption}

\newcommand{\owo}[1]{\textsc{OAgents}}

\definecolor{lightgreen}{RGB}{144, 238, 144} 
\definecolor{lightred}{RGB}{255, 105, 97}

\newtcolorbox{promptbox}[2][Prompt]{
colback=black!5!white,
arc=5pt, 
boxrule=0.5pt,
fonttitle=\bfseries,
title=#1, 
before upper={\small}, fontupper=\fontfamily{ptm}\selectfont,
colframe=#2, 
}
\definecolor{ogreen}{RGB}{34, 139, 34}

\usepackage{mathtools}
\usepackage{algorithm}
\usepackage[noend]{algpseudocode}
\usepackage{tabularx}
\usepackage{adjustbox}
\usepackage{float}
\usepackage{xspace}

\theoremstyle{plain}

\theoremstyle{definition}

\theoremstyle{remark}

\definecolor{demphcolor}{RGB}{144, 144, 144}

\definecolor{my_green}{RGB}{51,102,0}
\definecolor{my_red}{RGB}{204, 0, 0}

\newcommand{\modelname}{Hidden-Align\xspace}

\title{Right Makes Might: Aligning Verified Hidden States Empowers RL Reasoning}

\author[1,2*]{Ziyue Wang}
\author[1*]{Aomufei Yuan}
\author[2*]{Yongfu Zhu}
\author[2,3]{Shuai Dong}
\author[1,2]{Wenpu Liu}
\author[5]{Yiran Yao}
\author[1,2]{Weichu Xie}
\author[1,2]{Yuqi Xu}
\author[2,4]{Caoyuan Ma}
\author[2]{Wenqi Shao}
\author[2]{Xiaoying Zhang}
\author[2]{Nan Duan}
\author[2,\dagger]{Jiaqi Wang}

\affiliation[1]{Peking University}
\affiliation[2]{JINGDONG}
\affiliation[3]{Shanghai Innovation Institute}
\affiliation[4]{The University of Tokyo}
\affiliation[5]{Tianjin University}

\date{\today}

\correspondence{\email{wangziyue@tju.edu.cn}}

\abstract{
Reinforcement Learning from Verifiable Rewards (RLVR) has become the dominant approach for improving mathematical reasoning in large language models, yet current methods reduce each correct rollout to a single reward bit, ignoring the geometric structure shared among their hidden states.
Investigating this structure, we find that at the \textit{anchor token} (the position immediately before the answer marker), correct rollouts converge naturally because they must produce the same answer (cosine similarity ${\approx}\,0.84$), yet each retains residual variance from its unique reasoning path. Encouraging full alignment at this point pushes the model to extract a unified ``correct decision'' representation, reducing sensitivity to which reasoning path was taken.

Based on this observation, we propose \modelname, an auxiliary loss function that aligns the last-layer hidden states of correct rollouts at the anchor token during RL training, with zero overhead in both training and inference.
On eight mathematical reasoning benchmarks, \modelname improves average pass@$1$ over the DAPO baseline by 3.8, 6.2, and 5.4 percentage points on Qwen3-1.7B, 4B, and 14B respectively, with consistent pass@$k$ gains across all three scales, supported by ablations on loss type, anchor position, layer depth, and loss weight.
}

\begin{document}
\maketitle

\section{Introduction} \label{sec:intro}

Reinforcement Learning from Verifiable Rewards (RLVR) has become the dominant approach for mathematical reasoning in large language models~\cite{deepseek-r1,grpo,dapo,vapo,rewarding-unlikely}.
Methods such as DAPO and GRPO generate groups of rollouts per prompt, score each with a binary correctness reward, and update the policy via group-relative advantages.
This pipeline achieves strong gains on competition-level math benchmarks, yet it reduces every correct rollout to a single reward bit. When multiple rollouts in a group solve the same problem correctly, the geometric relationships among their internal representations are discarded. While regularization techniques such as KL penalties~\cite{instructgpt} exist, they operate on the output distribution. While recent work has used hidden states for reward prediction~\cite{reward-inside} and reward model regularization~\cite{regularizing-hs}, no existing method aligns correct rollouts' hidden states as a training signal during RL.

\begin{figure}[htbp]
    \centering
    \includegraphics[width=\textwidth]{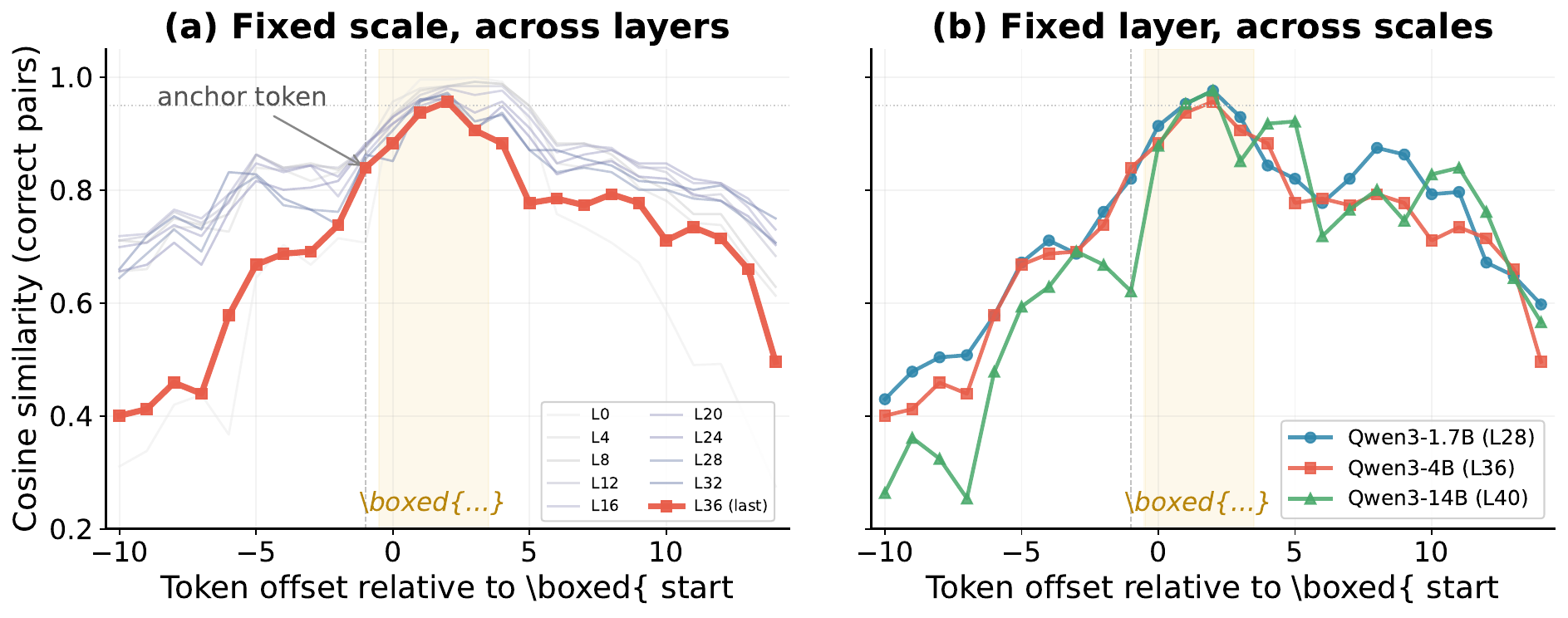}
    \caption{Pairwise cosine similarity of correct rollouts' hidden states, plotted against token position relative to \texttt{\textbackslash boxed\{}. \textbf{(a)}~Across all layers of Qwen3-4B, the last layer (L36, red) exhibits uniquely strong position-dependent variation, with a clear peak at the anchor token (offset $= -1$); shallower layers show weaker variation. \textbf{(b)}~At the last layer across three model scales (1.7B, 4B, 14B), the same pattern holds consistently: similarity peaks at the anchor token, forming a compact but not fully converged cluster ($\cos \approx 0.84$).}
    \label{fig:intro}
\end{figure}

We investigate whether this structure can be exploited as an auxiliary training signal by analyzing the hidden states of RL-trained models during reasoning. We begin by probing the pairwise cosine similarity of correct rollouts' hidden states across token positions~\cite{probing-llm,representation-engineering} (Figure~\ref{fig:intro}). The results show a clear positional structure: in the reasoning chain, hidden states are dispersed as each rollout follows a different path, reflecting the diverse reasoning strategies cultivated by RL training. Near the answer tokens, they converge sharply ($>0.9$) because all correct rollouts produce the same answer. Past the answer, similarity drops again. Between these regions sits the \textit{anchor token}, the position immediately before the answer marker \texttt{\textbackslash boxed\{}. Here, hidden states are compact (cosine similarity ${\approx}\,0.84$) yet not fully converged, occupying a unique sweet spot.

We also notice that the last layer's curve stands out from shallower layers, exhibiting far stronger position-dependent variation in Figure~\ref{fig:intro}(a). Repeating the analysis on the last layer across three model scales of Qwen3~\cite{qwen3}, we find the same pattern consistently in Figure~\ref{fig:intro}(b). Inspired by the effectiveness of representation alignment in other domains~\cite{repa}, we hypothesize that actively aligning these hidden states, encouraging their cosine similarity toward full agreement, would push the model to distill the common structure shared by diverse correct reasoning paths into a robust decision representation.

We propose \modelname, an auxiliary loss that maximizes the pairwise cosine similarity of correct rollouts' last-layer hidden states at the anchor token during RL training. Since the hidden states are already produced during rollout, this adds zero overhead in both training and inference. We systematically ablate each design choice (loss type, anchor position, layer depth, and loss weight), showing that the proposed configuration is uniquely effective.

In summary, we make the following contributions:
\begin{itemize}[leftmargin=*,nosep]
    \item We identify and quantify a consistent geometric phenomenon in RL-trained reasoning models: at the anchor token, correct rollouts' last-layer hidden states converge naturally but incompletely, and this pattern holds consistently across model scales.
    \item We propose \modelname, an auxiliary loss for RL training that aligns correct rollouts' hidden states at the anchor token, with a two-phase backward pass that enables exact gradient computation under micro-batch training (\textsection\ref{sec:method}).
    \item On eight math benchmarks across Qwen3-1.7B, 4B, and 14B, \modelname improves both pass@$1$ and pass@$k$ over the DAPO baseline, with zero training and inference overhead. Systematic ablations on loss type, position, layer, and weight confirm that the proposed configuration is uniquely effective (\textsection\ref{sec:exp}).
\end{itemize}

\section{Related Work}

\subsection{Reinforcement Learning for LLM Reasoning}

RLVR has become the standard approach for improving mathematical reasoning in LLMs.
DeepSeek-R1~\cite{deepseek-r1} demonstrated that pure RL training can induce emergent chain-of-thought reasoning, leading to broad adoption of group-relative optimization methods.
Among these, GRPO~\cite{grpo} computes advantages within each rollout group, eliminating the need for a separate value model, and DAPO~\cite{dapo} further refines it with clip-higher, dynamic sampling, and token-level loss rebalancing.
A growing family of variants further improves the policy optimization objective: Dr.~GRPO~\cite{drgrpo} corrects normalization biases, VAPO~\cite{vapo} stabilizes value estimation, and others address advantage shaping~\cite{gspo,gmpo,gfpo,gtpo}, negative-sample utilization~\cite{ngrpo,fapo}, KL penalty design~\cite{apo}, intrinsic exploration~\cite{prepo}, reward granularity~\cite{xie2026step}, error diversity~\cite{liu2026leveraging}, and rank bias~\cite{rewarding-unlikely}.
All these methods operate at the token-probability level, optimizing the policy gradient, advantage function, or sampling strategy.
\modelname is orthogonal: it introduces a training signal in the hidden representation space, and can be combined with any of the above algorithms.

\subsection{Representation Alignment as a Training Signal}

Aligning intermediate representations with a reference target is effective in several domains.
REPA~\cite{repa} aligns diffusion transformer hidden states to DINOv2 features via cosine similarity, achieving a 17.5$\times$ training speedup and state-of-the-art generation quality.
Earlier, contrastive and relational distillation methods~\cite{crd,rkd} established that aligning representational structure, rather than matching output distributions, transfers knowledge more effectively.
In the speech domain, \citet{closing-modality-gap} use a representation alignment reward during RL training to close the reasoning gap between text and speech modalities, demonstrating that hidden-state similarity can serve directly as a reward signal.
These works establish that representation alignment is a broadly effective training principle; we bring it to RL-based reasoning for the first time, aligning correct rollouts' hidden states at a specific token position rather than distilling from an external model.

\subsection{Hidden States in Reinforcement Learning}

Several recent works use LLM hidden states within the RL pipeline, though none perform alignment among correct rollouts.
\citet{reward-inside} show that a simple linear probe on hidden states can predict reward nearly as well as a full reward model, confirming that hidden states encode correctness information.
\citet{regularizing-hs} regularize the reward model's hidden states to prevent over-optimization, improving generalization.
\citet{reasoning-trajectories} reveal step-specific geometry in reasoning trajectories but do not intervene on these representations.
CRAFT~\cite{craft} uses contrastive learning to separate safe from unsafe reasoning trajectories, targeting safety rather than reasoning performance.
Representation-based intrinsic rewards in traditional RL~\cite{ride,foundation-novelty,lesr} are exploration-oriented, whereas \modelname is attraction-oriented: it rewards similarity among correct rollouts at a specific position, consolidating diverse reasoning paths into a unified decision representation.

\section{Methodology} \label{sec:method}

\begin{figure}[t]
    \centering
    \includegraphics[width=0.87\textwidth]{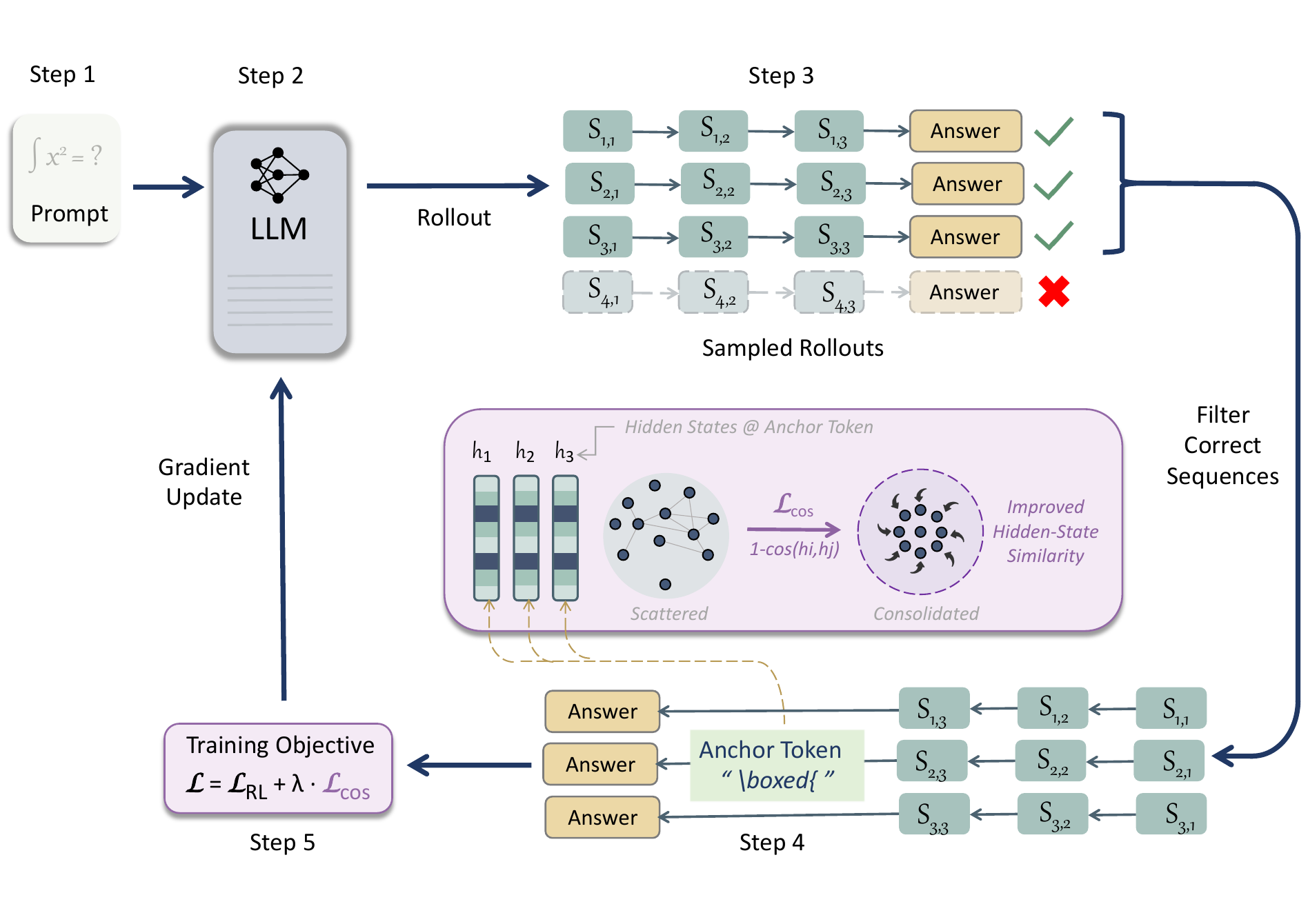}
    \caption{Overview of \modelname.\\
    \textbf{Steps 1--3}: A prompt is fed to the LLM, which generates multiple rollouts; correct sequences are identified by reward verification.\\
    \textbf{Step 4}: From correct sequences, we extract last-layer hidden states at the anchor token (immediately before \texttt{\textbackslash boxed\{}).\\
    \textbf{Center}: These hidden states are initially scattered; our alignment loss $\mathcal{L}_{\cos} = 1 - \cos(\mathbf{h}_i, \mathbf{h}_j)$ consolidates them.\\
    \textbf{Step 5}: The combined objective $\mathcal{L} = \mathcal{L}_{\text{DAPO}} + \lambda \cdot \mathcal{L}_{\cos}$ is used for gradient update.}
    \label{fig:method}
\end{figure}

Our method adds a single auxiliary loss to standard RLVR training.
Given a group of rollouts generated by the policy, we extract the last-layer hidden state at the \textit{anchor token} from each correct rollout and maximize their pairwise cosine similarity.
Figure~\ref{fig:method} illustrates the combined objective $\mathcal{L} = \mathcal{L}_{\text{DAPO}} + \lambda \cdot \mathcal{L}_{\cos}$ and the overall training pipeline.
We use DAPO~\cite{dapo} as our base RL algorithm throughout.
Below, we describe the base algorithm (\textsection\ref{sec:preliminary}), the alignment loss (\textsection\ref{sec:loss}), the gradient derivation (\textsection\ref{sec:backward}), the rationale for our design choices (\textsection\ref{sec:why}), and integration with the RL training pipeline (\textsection\ref{sec:impl}).

\subsection{Preliminary: Group-Relative Policy Optimization} \label{sec:preliminary}

We build upon DAPO~\cite{dapo}, a representative group-relative RLVR method.
For each training prompt $q$, the policy $\pi_\theta$ generates a group of $n$ responses $\{y_1, \ldots, y_n\}$.
Each response $y_i$ ($i = 1, \ldots, n$) receives a binary reward $r_i \in \{0, 1\}$ from a verifier.
The group-relative advantage is:
\begin{equation}
    A_i = \frac{r_i - \mu_r}{\sigma_r + \epsilon}
\end{equation}
where $\mu_r$ and $\sigma_r$ are the mean and standard deviation of rewards within the group, and $\epsilon$ is a small constant for numerical stability.
The policy is updated via clipped surrogate loss:
\begin{equation}
    \mathcal{L}_{\text{DAPO}} = -\mathbb{E}\left[\min\left(\rho_i A_i,\; \text{clip}(\rho_i, 1{-}\epsilon_l, 1{+}\epsilon_h) A_i\right)\right]
\end{equation}
where $\rho_i = \pi_\theta(y_i|q) / \pi_{\text{old}}(y_i|q)$ is the importance ratio, $\pi_{\text{old}}$ is the policy before the current update, and $\epsilon_l$, $\epsilon_h$ are the lower and upper clip bounds defined in DAPO.

\subsection{\modelname Loss} \label{sec:loss}

\paragraph{Anchor Token Definition.}
For each response $y_i$ in the group, we locate the first occurrence of the \texttt{\textbackslash boxed\{} token sequence in the generated output.
The \textit{anchor token} is defined as the token immediately preceding this marker, i.e., the point where the model has arrived at its answer but has not yet begun writing it.
We extract the hidden state $\mathbf{h}_i$ at this position from the last transformer layer, i.e., the final representation before the LM head projects to vocabulary logits.
As shown in Figure~\ref{fig:intro}, the anchor token occupies a middle ground in cosine similarity: higher than the reasoning chain (${\sim}0.4$) because correct rollouts are converging toward the same answer, but lower than the answer tokens themselves ($>0.9$), where similarity is too high for any auxiliary loss to provide meaningful gradients.

\paragraph{Alignment Loss.}
For each prompt group, let $\mathcal{C} = \{i : r_i = 1\}$ be the set of correct responses, and let $P = \{(i,j) : i,j \in \mathcal{C}, i < j\}$ be all correct-answer pairs.
We compute the average pairwise cosine similarity among correct rollouts' hidden states:
\begin{equation}
    \mathcal{L}_{\cos} = 1 - \frac{1}{|P|} \sum_{(i,j) \in P} \cos(\mathbf{h}_i, \mathbf{h}_j)
    \label{eq:cohesion}
\end{equation}
where $\cos(\mathbf{h}_i, \mathbf{h}_j) = \hat{\mathbf{h}}_i^\top \hat{\mathbf{h}}_j$ with $\hat{\mathbf{h}}_i = \mathbf{h}_i / \|\mathbf{h}_i\|$. The L2 normalization affects the gradient computation, which we derive in \textsection\ref{sec:backward}.
The loss is only computed when $|\mathcal{C}| \geq 2$; groups with fewer than two correct responses are skipped.

The total training loss combines the standard policy gradient with the alignment term:
\begin{equation}
    \mathcal{L} = \mathcal{L}_{\text{DAPO}} + \lambda \cdot \mathcal{L}_{\cos}, \quad \lambda = 0.001
    \label{eq:total}
\end{equation}

\paragraph{Design Choices.}
We consider five candidate loss types defined on the anchor token hidden states within each prompt group:
\begin{itemize}[leftmargin=*,nosep]
    \item \textbf{Alignment}: \textit{pull together} correct rollouts (maximize pairwise cosine similarity).
    \item \textbf{Separation}: \textit{push apart} correct and incorrect rollouts (minimize cross-group cosine similarity).
    \item \textbf{Negative}: \textit{push apart} incorrect rollouts from each other (minimize pairwise cosine similarity among incorrect).
    \item \textbf{Diversity}: \textit{push apart} all rollouts regardless of correctness (minimize global pairwise cosine similarity).
    \item \textbf{Cluster}: \textit{pull together} all rollouts regardless of correctness (maximize global pairwise cosine similarity).
\end{itemize}
We adopt alignment as our primary loss. Separation is likely redundant with the DAPO advantage; diversity and negative losses oppose the natural convergence; cluster conflates correct and incorrect rollouts. We also evaluate combinations of these losses.
The small $\lambda = 0.001$ reflects the fact that correct rollouts are already partially aligned at the anchor token ($\cos \approx 0.84$); a stronger signal would interfere with the primary RL objective. We verify all these choices in \textsection\ref{sec:ablation}.

\subsection{Gradient Derivation} \label{sec:backward}

\paragraph{Gradient of the Alignment Loss.}
We derive the gradient of $\mathcal{L}_{\cos}$ (Eq.~\ref{eq:cohesion}) with respect to the raw hidden state $\mathbf{h}_i$. Expanding the cosine similarity as $\cos(\mathbf{h}_i, \mathbf{h}_j) = \hat{\mathbf{h}}_i^\top \hat{\mathbf{h}}_j$ with $\hat{\mathbf{h}}_i = \mathbf{h}_i / \|\mathbf{h}_i\|$, we first compute the Jacobian of L2 normalization:
\begin{equation}
    \frac{\partial \hat{\mathbf{h}}_i}{\partial \mathbf{h}_i} = \frac{1}{\|\mathbf{h}_i\|} \left(\mathbf{I} - \hat{\mathbf{h}}_i \hat{\mathbf{h}}_i^\top \right)
\end{equation}
The gradient of a single cosine term $s_{ij} = \hat{\mathbf{h}}_i^\top \hat{\mathbf{h}}_j$ with respect to $\mathbf{h}_i$:
\begin{equation}
    \frac{\partial s_{ij}}{\partial \mathbf{h}_i} = \frac{1}{\|\mathbf{h}_i\|} \left(\hat{\mathbf{h}}_j - s_{ij} \, \hat{\mathbf{h}}_i \right)
\end{equation}
Therefore, the gradient of $\mathcal{L}_{\cos}$ with respect to $\mathbf{h}_i$ (for a correct sample $i$ in group $g$):
\begin{equation}
    \frac{\partial \mathcal{L}_{\cos}}{\partial \mathbf{h}_i} = -\frac{1}{|P|} \sum_{\substack{j \in \mathcal{C}_g \\ j \neq i}} \frac{1}{\|\mathbf{h}_i\|} \left(\hat{\mathbf{h}}_j - s_{ij} \, \hat{\mathbf{h}}_i \right)
    \label{eq:gradient}
\end{equation}
where $\mathcal{C}_g$ is the set of correct responses in group $g$.
Intuitively, the gradient pushes $\hat{\mathbf{h}}_i$ toward the other correct hidden states $\hat{\mathbf{h}}_j$ in its group, with a projection term $s_{ij} \hat{\mathbf{h}}_i$ that prevents trivial collapse to a single point.

\subsection{Why It Works} \label{sec:why}

Correct rollouts already cluster more tightly than correct--incorrect pairs at the anchor token, suggesting that the alignment loss reinforces an existing trend rather than imposing an artificial constraint. However, this clustering remains incomplete: correct rollouts arrive at the same answer via different reasoning paths, and their hidden states encode a mixture of path-specific details and a shared ``correct decision'' signal (see Appendix~\ref{app:cosine_dist} for the full distribution).

By aligning these hidden states, we do not suppress the diversity of reasoning paths, which unfolds earlier in the sequence and is unaffected by a loss applied only at the anchor token. Instead, we encourage the model to factor out path-specific variance at the point of decision, distilling the common structure that leads to a correct answer.

This explains the improvement on both pass@$1$ and pass@$k$, as confirmed in Figure~\ref{fig:pass_at_k}. A more consistent decision representation makes greedy decoding more likely to land on the correct answer (pass@$1$). Moreover, the process of distilling common structure across diverse paths deepens the model's understanding of what constitutes a correct solution, improving its ability to find correct answers even through novel reasoning paths (pass@$k$).

\subsection{Integration with RL Training Pipeline} \label{sec:impl}

\paragraph{Challenge: Micro-Batch Training.}
Computing Eq.~\ref{eq:gradient} requires access to all correct embeddings $\hat{\mathbf{h}}_j$ from the same group (i.e., rollouts generated from the same prompt).
In standard RL training, each optimization step processes a large mini-batch of rollouts, which is further divided into micro-batches to fit GPU memory (see Appendix~\ref{app:microbatch} for details). This structure is essential for stable gradient estimation and training efficiency.
However, samples from the same group may be split across different micro-batches, complicating the computation of the alignment loss across a complete group. Gathering all group members into memory simultaneously would cause out-of-memory failures under typical training configurations, while computing the loss on incomplete groups would yield approximate gradient estimates.

\paragraph{Two-Phase Solution.}
We decompose the backward pass into two phases:

\textbf{Phase 1 (Gather):} We forward-pass all micro-batches sequentially with gradient computation disabled, which avoids storing the computation graph and significantly reduces memory usage. From each forward pass, we collect the anchor embedding, group identifier, and correctness label for every sample. These embeddings are L2-normalized and detached from the computation graph to serve as fixed references in Phase 2.

\textbf{Phase 2 (Backward):} For each micro-batch chunk, we re-forward the correct samples in training mode to obtain gradient-attached embeddings $\hat{\mathbf{h}}_i$.
For each anchor $i$, we compute the loss against the detached references from the same group:
\begin{equation}
    \mathcal{L}_{\cos}^{(i)} = -\frac{1}{|P|} \sum_{\substack{j \in \mathcal{C}_g \\ j \neq i}} \hat{\mathbf{h}}_i^\top \bar{\hat{\mathbf{h}}}_j
\end{equation}
where $\bar{\hat{\mathbf{h}}}_j$ are detached (gradients flow only through $\hat{\mathbf{h}}_i$, not through $\bar{\hat{\mathbf{h}}}_j$).
The total loss $\mathcal{L}_{\cos} = \sum_i \mathcal{L}_{\cos}^{(i)}$ yields the exact same gradients as Eq.~\ref{eq:gradient}: although each term only backpropagates through one anchor, every pair $(i,j)$ is visited twice (once with $i$ as the anchor and once with $j$), so both sides receive the correct gradient.

This decomposition achieves exact gradient computation with bounded memory usage, and requires no modification to the standard mini-batch/micro-batch training structure.

\paragraph{Group-Preserving Batch Reordering.}
In data-parallel training, rollouts are distributed across GPUs. If samples from the same group are assigned to different GPUs, the alignment loss cannot access the complete set of correct pairs within that group. We address this by reordering the training batch so that all rollouts sharing the same prompt are co-located on a single GPU. To prevent memory imbalance, we additionally ensure that each GPU receives a similar total workload during reordering.

\section{Experiments} \label{sec:exp}

\subsection{Experimental Setup}

\paragraph{Models.}
We use Qwen3-4B as the primary model for main results and all ablation experiments, and Qwen3-1.7B and Qwen3-14B for scale verification.
We apply \modelname on top of DAPO training.

\paragraph{Training Data.}
We use DAPO-Math-17K~\cite{dapo}, a standard RLVR training set of 17,398 mathematical problems covering competition math, algebra, geometry, number theory, and combinatorics. All problems have answers enclosed in \texttt{\textbackslash boxed\{\}}, enabling automated reward verification.

\paragraph{Benchmarks.}
We evaluate on eight mathematical reasoning benchmarks spanning competition math and general math:
AIME 2024/2025/2026~\cite{aime}, AMC 2023/2024~\cite{amc}, HMMT Feb 2025~\cite{hmmt} (competition-level, 30--45 questions each);
Minerva Math~\cite{minerva} (general math), and OlympiadBench~\cite{olympiadbench} (olympiad-level).

\paragraph{Metrics.}
We report accuracy on all benchmarks following standard RLVR evaluation protocols.
Detailed evaluation settings (sampling parameters, number of trials) are provided in Appendix~\ref{app:training}.

\subsection{Main Results} \label{sec:main_results}

\begin{table}[htbp]
\centering
\small
\caption{Main results on mathematical reasoning benchmarks. Detailed evaluation settings are provided in the Appendix.}
\label{tab:main}
\resizebox{\textwidth}{!}{
\begin{tabular}{l c ccccccccc}
\toprule
\textbf{Method} & \textbf{Size} & \textbf{AIME24} & \textbf{AIME25} & \textbf{AIME26} & \textbf{AMC23} & \textbf{AMC24} & \textbf{HMMT} & \textbf{Minerva} & \textbf{Olympiad} & \textbf{Avg} \\
\midrule
Qwen3-4B (base) &  & 24.79 & 21.98 & 17.29 & 69.22 & 45.35 & 11.25 & 56.25 & 54.15 & 37.53 \\
\quad + DAPO & 4B & 48.12 & 43.44 & 40.31 & 88.28 & 58.75 & 19.79 & 53.68 & 63.35 & 51.97 \\
\quad + DAPO + \textbf{\modelname} &  & \textbf{59.17} & \textbf{51.46} & \textbf{50.31} & \textbf{91.41} & \textbf{62.15} & \textbf{27.08} & \textbf{56.62} & \textbf{67.06} & \textbf{58.16} \\
\midrule
Qwen3-1.7B (base) &  & 14.27 & 10.42 & 8.33 & 46.56 & 27.99 & 4.69 & 36.76 & 41.54 & 23.82 \\
\quad + DAPO & 1.7B & 29.06 & 25.83 & 19.90 & 68.12 & 47.43 & 14.06 & 44.85 & 49.70 & 37.37 \\
\quad + DAPO + \textbf{\modelname} &  & \textbf{31.87} & \textbf{27.50} & \textbf{24.27} & \textbf{73.83} & \textbf{47.57} & \textbf{16.98} & \textbf{52.21} & \textbf{55.34} & \textbf{41.20} \\
\midrule
Qwen3-14B (base) &  & 29.48 & 25.52 & 17.60 & 72.97 & 49.58 & 11.25 & 61.03 & 58.61 & 40.76 \\
\quad + DAPO & 14B & 60.94 & 51.98 & 49.79 & 92.27 & 64.10 & 31.35 & 58.82 & 72.85 & 60.26 \\
\quad + DAPO + \textbf{\modelname} &  & \textbf{68.12} & \textbf{61.67} & \textbf{60.94} & \textbf{93.98} & \textbf{65.62} & \textbf{34.58} & \textbf{64.71} & \textbf{75.82} & \textbf{65.68} \\
\bottomrule
\end{tabular}}
\end{table}

\begin{figure}[h!]
    \centering
    \includegraphics[width=0.75\textwidth]{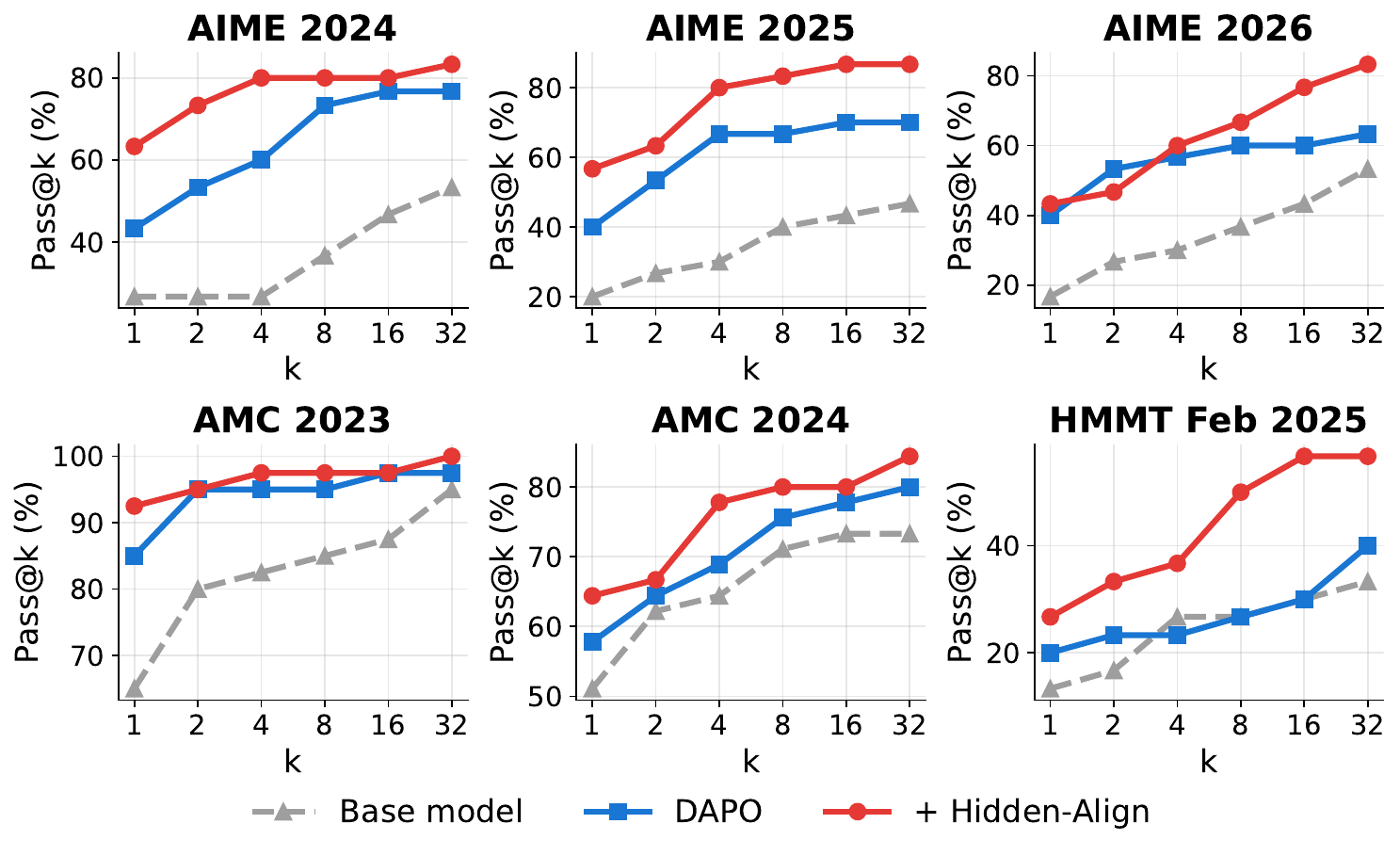}
    \caption{Pass@$k$ curves on six competition benchmarks for Qwen3-4B (temperature $0.2$, $n{=}32$). \modelname (red) outperforms DAPO (blue) and the base model (gray dashed) across most values of $k$.}
    \label{fig:pass_at_k}
\end{figure}

Table~\ref{tab:main} presents our main results.
On the 4B model, \modelname improves average accuracy by \textbf{6.19} percentage points over the DAPO baseline.
The largest gains are on AIME 2024 (+11.05pp), AIME 2026 (+10.00pp), and HMMT (+7.29pp).
The improvement is consistent across all three model scales, confirming that the method generalizes.

\paragraph{Pass@$k$ Analysis.}
Beyond greedy accuracy, we evaluate Pass@$k$ to measure the model's coverage of correct solutions across multiple attempts.
We sample $n=32$ responses per prompt at temperature $0.2$ and compute Pass@$k$ using the unbiased estimator from~\cite{chen2021evaluating}. Note that this differs from the greedy pass@$1$ in Table~\ref{tab:main}, which uses temperature $0$.
As shown in Figure~\ref{fig:pass_at_k}, \modelname achieves higher Pass@$k$ across most values of $k$ on all six competition benchmarks.
The gap is especially pronounced on harder benchmarks (AIME 2025/2026, HMMT).

\subsection{Ablation Studies} \label{sec:ablation}

We ablate each design choice of \modelname to verify its necessity.
All ablations are conducted on Qwen3-4B with DAPO training.

\begin{table}[htbp]
\centering
\small
\caption{Loss type ablation (all at anchor token, last layer, $\lambda=0.001$). Only alignment yields consistent improvement.}
\label{tab:loss_type}
\begin{tabular}{lccccc}
\toprule
\textbf{Loss Type} & \textbf{AIME24} & \textbf{AIME26} & \textbf{AMC24} & \textbf{HMMT} & \textbf{Avg} \\
\midrule
DAPO (baseline) & 48.12 & 40.31 & 58.75 & 19.79 & 41.74 \\
Alignment (ours) & \textbf{59.17} & \textbf{50.31} & \textbf{62.15} & \textbf{27.08} & \textbf{49.68} \\
Separation & 46.35 & 41.88 & 57.36 & 19.17 & 41.19 \\
Diversity & 49.69 & 45.21 & 61.94 & 24.17 & 45.25 \\
Negative & 49.06 & 45.83 & 58.40 & 24.38 & 44.42 \\
Cluster & 46.46 & 40.73 & 60.14 & 23.12 & 42.61 \\
Align. + Sep. & 45.62 & 41.25 & 59.93 & 22.40 & 42.30 \\
Align. + Neg. & 53.33 & 49.17 & 60.69 & 24.90 & 47.02 \\
Align. + Div. & 44.48 & 38.02 & 59.24 & 17.92 & 39.92 \\
\bottomrule
\end{tabular}
\end{table}

\paragraph{Loss Type.}
We compare the five loss types defined in \textsection\ref{sec:loss}.
Since correct rollouts already cluster at the anchor token, a pull-together signal (alignment) reinforces this existing trend.
In contrast, separation duplicates the role of the DAPO advantage, which already differentiates correct from incorrect via reward; diversity and negative losses actively fight the natural convergence; and cluster mixes correct and incorrect rollouts indiscriminately.
Table~\ref{tab:loss_type} confirms that only alignment yields consistent improvement, while combinations with other losses dilute or negate the gain.

\begin{figure}[htbp]
    \centering
    \includegraphics[width=\textwidth]{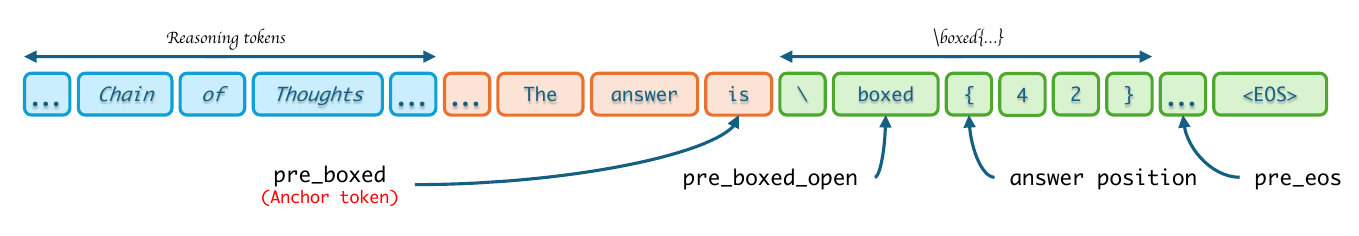}
    \caption{Token positions evaluated in the position ablation. The anchor token (pre\_boxed) is immediately before the \texttt{\textbackslash boxed\{} marker.}
    \label{fig:position_diagram}
\end{figure}

\begin{table}[htbp]
\centering
\small
\caption{Position ablation (alignment loss, last layer, $\lambda=0.001$).}
\label{tab:position}
\begin{tabular}{lccccc}
\toprule
\textbf{Anchor Position} & \textbf{AIME24} & \textbf{AIME26} & \textbf{AMC24} & \textbf{HMMT} & \textbf{Avg} \\
\midrule
DAPO (baseline) & 48.12 & 40.31 & 58.75 & 19.79 & 41.74 \\
pre\_boxed (ours) & \textbf{59.17} & \textbf{50.31} & \textbf{62.15} & \textbf{27.08} & \textbf{49.68} \\
pre\_eos & 43.65 & 41.04 & 57.92 & 21.35 & 40.99 \\
pre\_boxed\_open & 48.02 & 43.33 & 61.32 & 23.02 & 43.92 \\
answer position & 48.12 & 42.71 & 58.82 & 22.19 & 42.96 \\
\bottomrule
\end{tabular}
\end{table}

\paragraph{Anchor Position.}
We apply the alignment loss at different token positions relative to \texttt{\textbackslash boxed\{}.
Figure~\ref{fig:position_diagram} illustrates the four candidate positions. As discussed in \textsection\ref{sec:loss}, the anchor token occupies a unique sweet spot in cosine similarity; other positions are either too converged (answer tokens) or too dispersed (reasoning chain). Table~\ref{tab:position} confirms that only the anchor token yields substantial improvement.

\begin{table}[htbp]
\centering
\small
\caption{Layer ablation (alignment loss, anchor token, $\lambda=0.001$).}
\label{tab:layer}
\begin{tabular}{lccccc}
\toprule
\textbf{Layer} & \textbf{AIME24} & \textbf{AIME26} & \textbf{AMC24} & \textbf{HMMT} & \textbf{Avg} \\
\midrule
DAPO (baseline) & 48.12 & 40.31 & 58.75 & 19.79 & 41.74 \\
Layer 16 & 45.00 & 43.85 & 62.15 & 24.38 & 43.85 \\
Layer 24 & 47.40 & 44.27 & 61.88 & 21.98 & 43.88 \\
Layer 28 & 44.69 & 44.58 & 62.99 & 20.62 & 43.22 \\
Layer 32 & 49.48 & 43.44 & 60.07 & 23.75 & 44.19 \\
Layer 36 (ours) & \textbf{59.17} & \textbf{50.31} & \textbf{62.15} & \textbf{27.08} & \textbf{49.68} \\
\bottomrule
\end{tabular}
\end{table}

\paragraph{Layer Depth.}
We apply the alignment loss at different transformer layers. As shown in Figure~\ref{fig:intro}(a), only the last layer exhibits strong position-dependent variation in cosine similarity; shallower layers show weaker variation, meaning the anchor token is less distinctive at those depths. Table~\ref{tab:layer} confirms that only the last layer (Layer 36) produces a clear gain, while intermediate layers show marginal or no improvement.

\begin{table}[htbp]
\centering
\small
\caption{Loss weight ablation (alignment loss, anchor token, last layer).}
\label{tab:lambda}
\begin{tabular}{lccccc}
\toprule
\textbf{$\lambda$} & \textbf{AIME24} & \textbf{AIME26} & \textbf{AMC24} & \textbf{HMMT} & \textbf{Avg} \\
\midrule
DAPO (baseline) & 48.12 & 40.31 & 58.75 & 19.79 & 41.74 \\
0.0001 & 47.29 & 44.90 & 59.03 & 20.42 & 42.91 \\
0.0005 & 50.21 & 43.75 & 59.31 & 21.77 & 43.76 \\
\textbf{0.001 (ours)} & \textbf{59.17} & \textbf{50.31} & \textbf{62.15} & \textbf{27.08} & \textbf{49.68} \\
0.005 & 42.71 & 39.06 & 57.15 & 22.81 & 40.43 \\
0.01$^\dagger$ & 35.83 & 35.83 & 52.71 & 19.06 & 35.86 \\
\bottomrule
\end{tabular}
\par\smallskip
{\footnotesize $^\dagger$ Training diverged after step 100; results reported at step 80.}
\end{table}

\paragraph{Loss Weight.}
We vary $\lambda$ from $0.0001$ to $0.01$. Since correct rollouts are already partially aligned at the anchor token ($\cos \approx 0.84$), only a moderate $\lambda$ is needed. Too small a value provides insufficient signal; too large a value overpowers the RL objective and destabilizes training. Table~\ref{tab:lambda} shows that $\lambda = 0.001$ is optimal, with $\lambda = 0.01$ causing training divergence.

\subsection{Analysis} \label{sec:analysis}

\begin{figure}[htbp]
    \centering
    \includegraphics[width=0.7\textwidth]{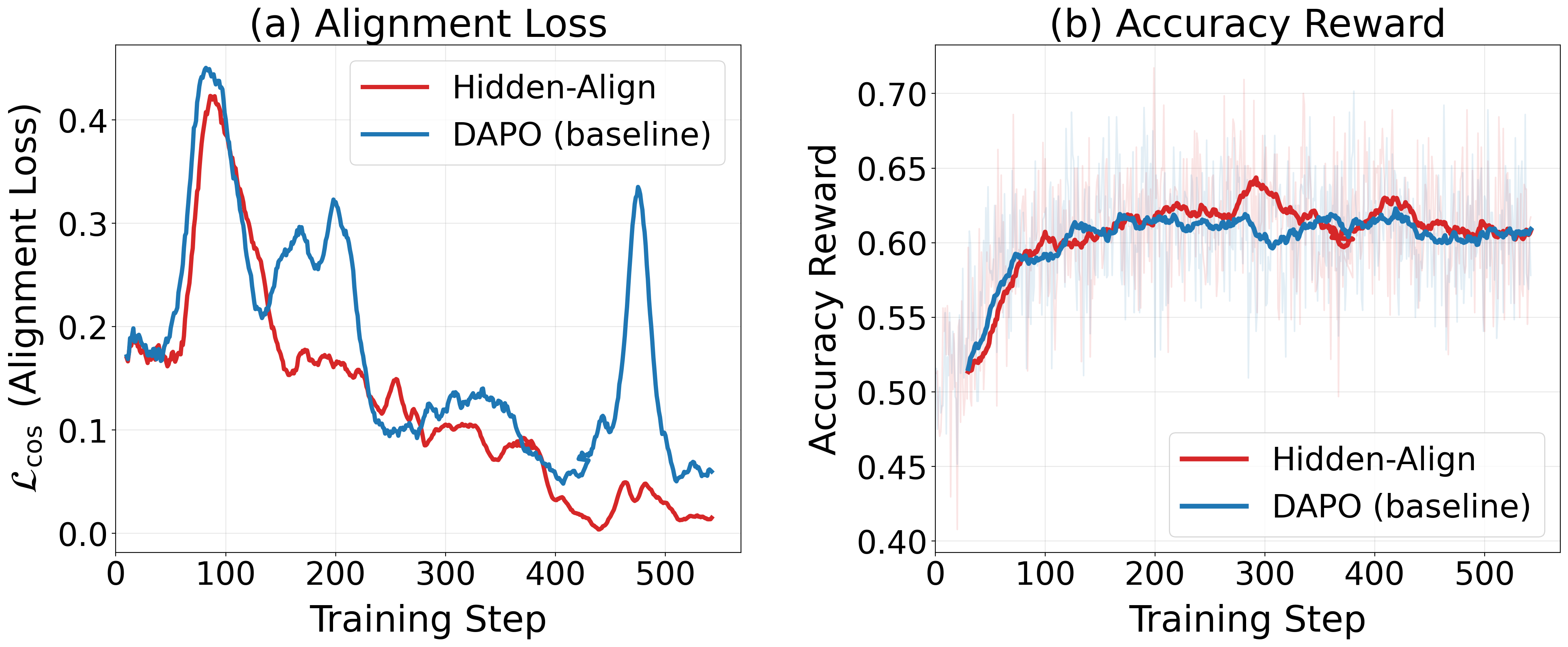}
    \caption{(a)~Alignment loss $\mathcal{L}_{\cos}$ and (b)~accuracy reward over training steps for Qwen3-4B. Hidden-Align (red) shows active alignment loss reduction; the baseline's alignment loss (monitored but not optimized) spikes in later stages. In (b), faint lines show raw per-step values and bold lines show smoothed trends. Hidden-Align generally achieves higher reward.}
    \label{fig:pos_loss_acc}
\end{figure}

\paragraph{Alignment Loss and Reward.}
Figure~\ref{fig:pos_loss_acc} compares the alignment loss $\mathcal{L}_{\cos}$ and accuracy reward between Hidden-Align and the DAPO baseline over the course of training.
In (a), the alignment loss for Hidden-Align starts high and gradually decreases, indicating that correct rollouts' hidden states are being actively aligned at the anchor token. For comparison, we also monitor the same metric on the DAPO baseline (which does not optimize it): the baseline shows a similar but weaker downward trend in early training, suggesting that RL training alone induces partial alignment. Moreover, around step 400--500, the baseline's alignment loss spikes sharply, indicating that the natural clustering breaks down in later training stages. In contrast, Hidden-Align maintains stable alignment throughout.
In (b), Hidden-Align generally achieves higher accuracy reward than the baseline, particularly during steps 200--300. The overall trend confirms that aligning correct rollouts' hidden states provides a beneficial training signal.

\begin{figure}[h]
    \centering
    \includegraphics[width=0.7\textwidth]{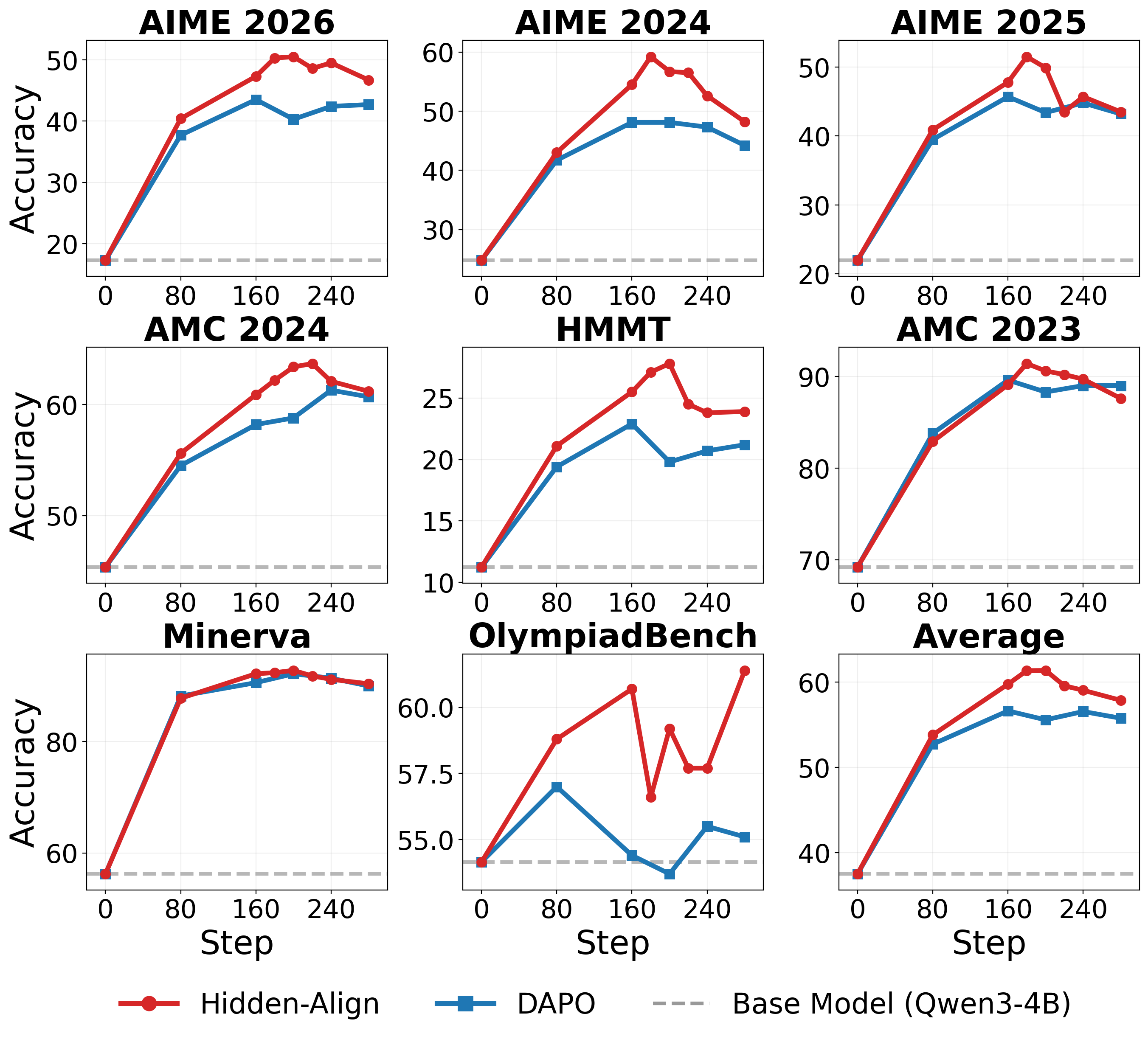}
    \caption{Per-benchmark accuracy over training steps for Qwen3-4B. Hidden-Align (red) outperforms DAPO (blue) on most benchmarks throughout training, with the largest gains on competition-level benchmarks (AIME, HMMT). The gray dashed line indicates the base model (before RL training). Both methods peak around step 160--200 and gradually decline afterward.}
    \label{fig:acc_steps}
\end{figure}

\paragraph{Accuracy over Training Steps.}
Figure~\ref{fig:acc_steps} shows per-benchmark accuracy at multiple training checkpoints.
Hidden-Align consistently outperforms DAPO on competition-level benchmarks (AIME 2024/2025/2026, HMMT), with the gap emerging as early as step 80 and widening through step 160--200.
On easier benchmarks (AMC 2023/2024), both methods perform similarly, as they are closer to the ceiling.
Both methods peak around step 160--200 and gradually decline afterward; we report the best checkpoint for each configuration.

\section{Conclusion}

We present \modelname, a lightweight auxiliary loss that aligns correct rollouts' hidden states at the anchor token during RL training.
Starting from the observation that correct rollouts naturally cluster at this position but retain residual variance from diverse reasoning paths, we show that a single alignment loss ($\lambda = 0.001$) at the last transformer layer can distill their shared decision structure into a more robust representation.
On eight mathematical reasoning benchmarks, \modelname improves average accuracy over the DAPO baseline by 3.83, 6.19, and 5.42 percentage points on Qwen3-1.7B, 4B, and 14B respectively, with consistent pass@$k$ gains and zero overhead in both training and inference.
Systematic ablations on loss type, anchor position, layer depth, and loss weight confirm that the proposed configuration is uniquely effective.

\newpage

\section*{Limitations}

\begin{itemize}[leftmargin=*,nosep]
    \item We validate \modelname only on mathematical reasoning tasks with \texttt{\textbackslash boxed\{\}} answer format. Extending to other tasks requires identifying the corresponding anchor position for each answer format.
    \item The hyperparameter $\lambda$ is fixed throughout training. Adaptive scheduling conditioned on training progress could further improve results.
    \item We verify on 1.7B, 4B, and 14B models. Validation on even larger scales (70B+) is future work.
\end{itemize}

\section*{Ethical Considerations}

This work improves mathematical reasoning in LLMs through a representation-level auxiliary loss applied during RL training. It does not involve human subjects, private data, or content generation in sensitive domains. The training data (DAPO-Math-17K) consists of publicly available mathematical problems. We do not foresee specific ethical risks beyond those common to general-purpose LLM reasoning research.

\newpage

\bibliographystyle{plainnat}
\bibliography{main}

\begin{thebibliography}{37}
\providecommand{\natexlab}[1]{#1}
\providecommand{\url}[1]{\texttt{#1}}
\expandafter\ifx\csname urlstyle\endcsname\relax
  \providecommand{\doi}[1]{doi: #1}\else
  \providecommand{\doi}{doi: \begingroup \urlstyle{rm}\Url}\fi

\bibitem[Belinkov(2022)]{probing-llm}
Yonatan Belinkov.
\newblock Probing classifiers: Promises, shortcomings, and advances.
\newblock \emph{Computational Linguistics}, 48\penalty0 (1):\penalty0 207--219, 2022.

\bibitem[Chen et~al.(2021)Chen, Tworek, Jun, Yuan, Pinto, Kaplan, Edwards, Burda, Joseph, Brockman, et~al.]{chen2021evaluating}
Mark Chen, Jerry Tworek, Heewoo Jun, Qiming Yuan, Henrique Ponde de~Oliveira Pinto, Jared Kaplan, Harri Edwards, Yuri Burda, Nicholas Joseph, Greg Brockman, et~al.
\newblock Evaluating large language models trained on code.
\newblock \emph{arXiv preprint arXiv:2107.03374}, 2021.

\bibitem[Ding et~al.(2025)Ding, Zhang, Li, Lin, and Zhang]{fapo}
Yuyang Ding, Chi Zhang, Juntao Li, Haibin Lin, and Min Zhang.
\newblock Fapo: flawed-aware policy optimization for efficient and reliable reasoning.
\newblock \emph{arXiv preprint arXiv:2510.22543}, 2025.

\bibitem[Guo et~al.(2025{\natexlab{a}})Guo, Yang, Zhang, Song, Wang, Zhu, Xu, Zhang, Ma, Bi, et~al.]{deepseek-r1}
Daya Guo, Dejian Yang, Haowei Zhang, Junxiao Song, Peiyi Wang, Qihao Zhu, Runxin Xu, Ruoyu Zhang, Shirong Ma, Xiao Bi, et~al.
\newblock Deepseek-r1: Incentivizing reasoning capability in llms via reinforcement learning.
\newblock \emph{arXiv preprint arXiv:2501.12948}, 2025{\natexlab{a}}.

\bibitem[Guo et~al.(2025{\natexlab{b}})Guo, Wu, and Philip]{reward-inside}
Jizhou Guo, Zhaomin Wu, and S~Yu Philip.
\newblock Reward inside the model: A lightweight hidden-state reward model for llm's best-of-n sampling.
\newblock In \emph{2nd AI for Math Workshop@ ICML 2025}, 2025{\natexlab{b}}.

\bibitem[Gupta et~al.(2022)Gupta, Karkus, Che, Xu, and Pavone]{foundation-novelty}
Tarun Gupta, Peter Karkus, Tong Che, Danfei Xu, and Marco Pavone.
\newblock Foundation models for semantic novelty in reinforcement learning.
\newblock \emph{arXiv preprint arXiv:2211.04878}, 2022.

\bibitem[{Harvard-MIT Mathematics Tournament}(2025)]{hmmt}
{Harvard-MIT Mathematics Tournament}.
\newblock {HMMT} february competition.
\newblock \url{https://www.hmmt.org/}, 2025.

\bibitem[He et~al.(2025)He, Fried, and Welleck]{rewarding-unlikely}
Andre~Wang He, Daniel Fried, and Sean Welleck.
\newblock Rewarding the unlikely: Lifting grpo beyond distribution sharpening.
\newblock In \emph{Proceedings of the 2025 Conference on Empirical Methods in Natural Language Processing}, pages 25559--25571, 2025.

\bibitem[He et~al.(2024)He, Luo, Bai, Hu, Thai, Shen, Hu, Han, Huang, Zhang, et~al.]{olympiadbench}
Chaoqun He, Renjie Luo, Yuzhuo Bai, Shengding Hu, Zhen Thai, Junhao Shen, Jinyi Hu, Xu~Han, Yujie Huang, Yuxiang Zhang, et~al.
\newblock Olympiadbench: A challenging benchmark for promoting agi with olympiad-level bilingual multimodal scientific problems.
\newblock In \emph{Proceedings of the 62nd Annual Meeting of the Association for Computational Linguistics (Volume 1: Long Papers)}, pages 3828--3850, 2024.

\bibitem[Hong et~al.(2025)Hong, Guo, Xia, Wang, Zhang, Jin, and Zhao]{apo}
Minjie Hong, Zirun Guo, Yan Xia, Zehan Wang, Ziang Zhang, Tao Jin, and Zhou Zhao.
\newblock Apo: Enhancing reasoning ability of mllms via asymmetric policy optimization.
\newblock \emph{arXiv preprint arXiv:2506.21655}, 2025.

\bibitem[Lewkowycz et~al.(2022)Lewkowycz, Andreassen, Dohan, Dyer, Michalewski, Ramasesh, Slone, Anil, Schlag, Gutman-Solo, et~al.]{minerva}
Aitor Lewkowycz, Anders Andreassen, David Dohan, Ethan Dyer, Henryk Michalewski, Vinay Ramasesh, Ambrose Slone, Cem Anil, Imanol Schlag, Theo Gutman-Solo, et~al.
\newblock Solving quantitative reasoning problems with language models.
\newblock \emph{Advances in neural information processing systems}, 35:\penalty0 3843--3857, 2022.

\bibitem[Liu et~al.(2026)Liu, Xu, Xie, Zhu, Dong, Wang, Shao, Zhang, Yang, Duan, et~al.]{liu2026leveraging}
Wenpu Liu, Yuqi Xu, Weichu Xie, Yongfu Zhu, Shuai Dong, Ziyue Wang, Wenqi Shao, Xiaoying Zhang, Tong Yang, Nan Duan, et~al.
\newblock Leveraging error diversity in group rollouts for reinforcement learning.
\newblock \emph{arXiv preprint arXiv:2605.17333}, 2026.

\bibitem[Liu et~al.(2025)Liu, Chen, Li, Qi, Pang, Du, Lee, and Lin]{drgrpo}
Zichen Liu, Changyu Chen, Wenjun Li, Penghui Qi, Tianyu Pang, Chao Du, Wee~Sun Lee, and Min Lin.
\newblock Understanding r1-zero-like training: A critical perspective.
\newblock \emph{arXiv preprint arXiv:2503.20783}, 2025.

\bibitem[Luo et~al.(2026)Luo, Wang, Yu, Wang, and Chen]{craft}
Haozheng Luo, Yimin Wang, Jiahao Yu, Binghui Wang, and Yan Chen.
\newblock Contrastive reasoning alignment: Reinforcement learning from hidden representations.
\newblock \emph{arXiv preprint arXiv:2603.17305}, 2026.

\bibitem[{Mathematical Association of America}(2024{\natexlab{a}})]{aime}
{Mathematical Association of America}.
\newblock {AIME} problems and solutions.
\newblock \url{https://maa.org/}, 2024{\natexlab{a}}.

\bibitem[{Mathematical Association of America}(2024{\natexlab{b}})]{amc}
{Mathematical Association of America}.
\newblock {AMC} 10/12 problems and solutions.
\newblock \url{https://maa.org/}, 2024{\natexlab{b}}.

\bibitem[Nan et~al.(2025)Nan, Chen, Huang, Lu, Wang, Xie, Xiong, Zeng, Zhou, Li, et~al.]{ngrpo}
Gongrui Nan, Siye Chen, Jing Huang, Mengyu Lu, Dexun Wang, Chunmei Xie, Weiqi Xiong, Xianzhou Zeng, Qixuan Zhou, Yadong Li, et~al.
\newblock Ngrpo: Negative-enhanced group relative policy optimization.
\newblock \emph{arXiv preprint arXiv:2509.18851}, 2025.

\bibitem[Ouyang et~al.(2022)Ouyang, Wu, Jiang, Almeida, Wainwright, Mishkin, Zhang, Agarwal, Slama, Ray, et~al.]{instructgpt}
Long Ouyang, Jeffrey Wu, Xu~Jiang, Diogo Almeida, Carroll Wainwright, Pamela Mishkin, Chong Zhang, Sandhini Agarwal, Katarina Slama, Alex Ray, et~al.
\newblock Training language models to follow instructions with human feedback.
\newblock \emph{Advances in neural information processing systems}, 35:\penalty0 27730--27744, 2022.

\bibitem[Park et~al.(2019)Park, Kim, Lu, and Cho]{rkd}
Wonpyo Park, Dongju Kim, Yan Lu, and Minsu Cho.
\newblock Relational knowledge distillation.
\newblock In \emph{Proceedings of the IEEE/CVF conference on computer vision and pattern recognition}, pages 3967--3976, 2019.

\bibitem[Raileanu and Rockt{\"a}schel(2020)]{ride}
Roberta Raileanu and Tim Rockt{\"a}schel.
\newblock Ride: Rewarding impact-driven exploration for procedurally-generated environments.
\newblock \emph{arXiv preprint arXiv:2002.12292}, 2020.

\bibitem[Shao et~al.(2024)Shao, Wang, Zhu, Xu, Song, Bi, Zhang, Zhang, Li, Wu, et~al.]{grpo}
Zhihong Shao, Peiyi Wang, Qihao Zhu, Runxin Xu, Junxiao Song, Xiao Bi, Haowei Zhang, Mingchuan Zhang, YK~Li, Yang Wu, et~al.
\newblock Deepseekmath: Pushing the limits of mathematical reasoning in open language models.
\newblock \emph{arXiv preprint arXiv:2402.03300}, 2024.

\bibitem[Shrivastava et~al.(2025)Shrivastava, Awadallah, Balachandran, Garg, Behl, and Papailiopoulos]{gfpo}
Vaishnavi Shrivastava, Ahmed Awadallah, Vidhisha Balachandran, Shivam Garg, Harkirat Behl, and Dimitris Papailiopoulos.
\newblock Sample more to think less: Group filtered policy optimization for concise reasoning.
\newblock \emph{arXiv preprint arXiv:2508.09726}, 2025.

\bibitem[Simoni et~al.(2025)Simoni, Fontana, Rossolini, Saracino, and Mori]{gtpo}
Marco Simoni, Aleksandar Fontana, Giulio Rossolini, Andrea Saracino, and Paolo Mori.
\newblock Gtpo: Stabilizing group relative policy optimization via gradient and entropy control.
\newblock \emph{arXiv preprint arXiv:2508.03772}, 2025.

\bibitem[Sun et~al.(2026)Sun, Dong, Qiao, Lin, Zhang, and Rajmohan]{reasoning-trajectories}
Lihao Sun, Hang Dong, Bo~Qiao, Qingwei Lin, Dongmei Zhang, and Saravan Rajmohan.
\newblock Llm reasoning as trajectories: Step-specific representation geometry and correctness signals.
\newblock \emph{arXiv preprint arXiv:2604.05655}, 2026.

\bibitem[Sun et~al.(2025)Sun, Guo, Kok, Wang, Wen, and Zhang]{prepo}
Yan Sun, Jia Guo, Stanley Kok, Zihao Wang, Zujie Wen, and Zhiqiang Zhang.
\newblock Efficient reinforcement learning for large language models with intrinsic exploration.
\newblock \emph{arXiv preprint arXiv:2511.00794}, 2025.

\bibitem[Tian et~al.(2019)Tian, Krishnan, and Isola]{crd}
Yonglong Tian, Dilip Krishnan, and Phillip Isola.
\newblock Contrastive representation distillation.
\newblock \emph{arXiv preprint arXiv:1910.10699}, 2019.

\bibitem[Wang et~al.(2024)Wang, Qu, Jiang, Shao, Liu, Yang, and Ji]{lesr}
Boyuan Wang, Yun Qu, Yuhang Jiang, Jianzhun Shao, Chang Liu, Wenming Yang, and Xiangyang Ji.
\newblock Llm-empowered state representation for reinforcement learning.
\newblock \emph{arXiv preprint arXiv:2407.13237}, 2024.

\bibitem[Wang et~al.(2026)Wang, Lu, Zhang, Liu, Lu, Li, and Wu]{closing-modality-gap}
Chaoren Wang, Heng Lu, Xueyao Zhang, Shujie Liu, Yan Lu, Jinyu Li, and Zhizheng Wu.
\newblock Closing the modality reasoning gap for speech large language models.
\newblock \emph{arXiv preprint arXiv:2601.05543}, 2026.

\bibitem[Xie et~al.(2026)Xie, Zhao, Liu, Zhu, Chen, Ye, Chen, Xu, Dong, Wang, et~al.]{xie2026step}
Weichu Xie, Haozhe Zhao, Wenpu Liu, Yongfu Zhu, Liang Chen, Minghao Ye, Zirong Chen, Yuqi Xu, Shuai Dong, Ziyue Wang, et~al.
\newblock Step-wise rubric rewards for llm reasoning.
\newblock \emph{arXiv preprint arXiv:2605.17291}, 2026.

\bibitem[Yang et~al.(2025)Yang, Li, Yang, Zhang, Hui, Zheng, Yu, Gao, Huang, Lv, et~al.]{qwen3}
An~Yang, Anfeng Li, Baosong Yang, Beichen Zhang, Binyuan Hui, Bo~Zheng, Bowen Yu, Chang Gao, Chengen Huang, Chenxu Lv, et~al.
\newblock Qwen3 technical report.
\newblock \emph{arXiv preprint arXiv:2505.09388}, 2025.

\bibitem[Yang et~al.(2024)Yang, Ding, Lin, Zhang, and Zhang]{regularizing-hs}
Rui Yang, Ruomeng Ding, Yong Lin, Huan Zhang, and Tong Zhang.
\newblock Regularizing hidden states enables learning generalizable reward model for llms.
\newblock \emph{Advances in Neural Information Processing Systems}, 37:\penalty0 62279--62309, 2024.

\bibitem[Yu et~al.(2026)Yu, Zhang, Zhu, Yuan, Zuo, Yue, Dai, Fan, Liu, Liu, et~al.]{dapo}
Qiying Yu, Zheng Zhang, Ruofei Zhu, Yufeng Yuan, Xiaochen Zuo, Yu~Yue, Weinan Dai, Tiantian Fan, Gaohong Liu, Lingjun Liu, et~al.
\newblock Dapo: An open-source llm reinforcement learning system at scale.
\newblock \emph{Advances in Neural Information Processing Systems}, 38:\penalty0 113222--113244, 2026.

\bibitem[Yu et~al.(2024)Yu, Kwak, Jang, Jeong, Huang, Shin, and Xie]{repa}
Sihyun Yu, Sangkyung Kwak, Huiwon Jang, Jongheon Jeong, Jonathan Huang, Jinwoo Shin, and Saining Xie.
\newblock Representation alignment for generation: Training diffusion transformers is easier than you think.
\newblock \emph{arXiv preprint arXiv:2410.06940}, 2024.

\bibitem[Yue et~al.(2025)Yue, Yuan, Yu, Zuo, Zhu, Xu, Chen, Wang, Fan, Du, et~al.]{vapo}
Yu~Yue, Yufeng Yuan, Qiying Yu, Xiaochen Zuo, Ruofei Zhu, Wenyuan Xu, Jiaze Chen, Chengyi Wang, TianTian Fan, Zhengyin Du, et~al.
\newblock Vapo: Efficient and reliable reinforcement learning for advanced reasoning tasks.
\newblock \emph{arXiv preprint arXiv:2504.05118}, 2025.

\bibitem[Zhao et~al.(2025)Zhao, Liu, Liu, Chen, Wu, Hao, Lv, Huang, Cui, Ye, et~al.]{gmpo}
Yuzhong Zhao, Yue Liu, Junpeng Liu, Jingye Chen, Xun Wu, Yaru Hao, Tengchao Lv, Shaohan Huang, Lei Cui, Qixiang Ye, et~al.
\newblock Geometric-mean policy optimization.
\newblock \emph{arXiv preprint arXiv:2507.20673}, 2025.

\bibitem[Zheng et~al.(2025)Zheng, Liu, Li, Chen, Yu, Gao, Dang, Liu, Men, Yang, et~al.]{gspo}
Chujie Zheng, Shixuan Liu, Mingze Li, Xiong-Hui Chen, Bowen Yu, Chang Gao, Kai Dang, Yuqiong Liu, Rui Men, An~Yang, et~al.
\newblock Group sequence policy optimization.
\newblock \emph{arXiv preprint arXiv:2507.18071}, 2025.

\bibitem[Zou et~al.(2023)Zou, Phan, Chen, Campbell, Guo, Ren, Pan, Yin, Mazeika, Dombrowski, et~al.]{representation-engineering}
Andy Zou, Long Phan, Sarah Chen, James Campbell, Phillip Guo, Richard Ren, Alexander Pan, Xuwang Yin, Mantas Mazeika, Ann-Kathrin Dombrowski, et~al.
\newblock Representation engineering: A top-down approach to ai transparency.
\newblock \emph{arXiv preprint arXiv:2310.01405}, 2023.

\end{thebibliography}

\newpage
\appendix

\begin{center}
{\Large \textbf{APPENDIX}}
\end{center}
\vspace{1em}

\section{Cosine Similarity Distributions}
\label{app:cosine_dist}

To verify that correct rollouts cluster more tightly than other pairs at the anchor token, we compute pairwise cosine similarities on 1,000 prompts from the DAPO-Math-17K training set, using the base Qwen3-4B model with 10 rollouts per prompt.

\begin{figure}[htbp]
    \centering
    \includegraphics[width=\textwidth]{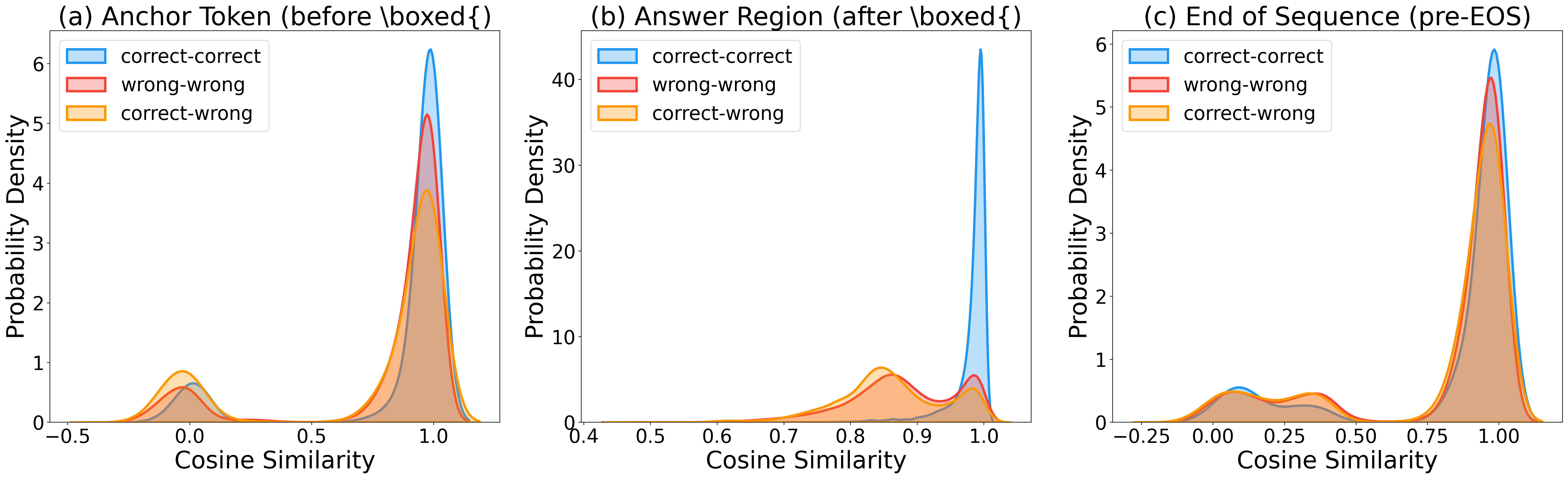}
    \caption{Pairwise cosine similarity distributions at three token positions (Qwen3-4B, Layer 36). (a)~At the anchor token, correct--correct pairs cluster more tightly than other pairs. (b)~In the answer region, all correct-correct pairs converge near 1.0. (c)~At the end of sequence, distributions overlap with no useful structure.}
    \label{fig:cosine_dist}
\end{figure}

Figure~\ref{fig:cosine_dist} shows the distribution of pairwise cosine similarity at the last layer (Layer 36) for three token positions.
Position~(a) is the \textit{anchor token}, the token immediately before the \texttt{\textbackslash boxed\{} marker, where the model has finished reasoning but has not yet begun writing the answer. Here, correct--correct pairs (blue) are clearly shifted toward higher similarity compared to correct--wrong (orange) and wrong--wrong (red) pairs, confirming that correct rollouts naturally cluster more tightly at this position.
Position~(b) is the token at the start of the answer region, where the model is about to generate the answer digits. All three distributions collapse to near 1.0, and this convergence is dominated by correct--correct pairs, since correct rollouts produce the same answer and thus their hidden states become nearly identical. This leaves no room for an auxiliary loss to provide meaningful gradient signal.
Position~(c) is the last token of the response (pre-EOS), where the model is about to terminate generation. While some clustering is visible, this position carries no semantic significance for reasoning; any similarity here reflects formatting patterns rather than decision quality.

\section{Training Details}
\label{app:training}

\paragraph{Mini-Batch and Micro-Batch in RL Training.}
\label{app:microbatch}
In RLVR training, each optimization step begins with a \textit{rollout phase}: the current policy generates a large batch of responses (the mini-batch) across many prompts. This mini-batch must be large enough to provide stable gradient estimates, since the group-relative advantage $A_i$ is computed within each prompt group and benefits from seeing many groups per update. However, the subsequent \textit{training phase}---which runs forward and backward passes through the full model---cannot fit the entire mini-batch into GPU memory at once. The mini-batch is therefore split into smaller \textit{micro-batches} that are processed sequentially, with gradients accumulated before each optimizer step. This mini-batch/micro-batch structure is standard in RL training and cannot be easily modified without affecting training stability or efficiency.

\begin{table}[h]
\centering
\small
\caption{Training hyperparameters.}
\begin{tabular}{ll}
\toprule
\textbf{Parameter} & \textbf{Value} \\
\midrule
Base model & Qwen3-4B \\
 & Qwen3-1.7B \\
 & Qwen3-14B \\
Training data & DAPO-Math-17K \\
RL algorithm & DAPO \\
Learning rate & $1 \times 10^{-6}$ \\
Optimizer & AdamW ($\beta_1{=}0.9$, $\beta_2{=}0.999$) \\
Weight decay & $1 \times 10^{-2}$ \\
LR warmup & 0 \\
Clip ratio (high/low) & 0.2 / 0.2 \\
KL penalty & disabled \\
Online filtering & enabled \\
Rollout $n$ & 10 \\
Rollout batch size & 64 \\
Global batch size & 64 \\
Micro batch size & 1 \\
Max prompt length & 2048 \\
Max response length & 8192 \\
Tensor parallel size & 2 \\
GPUs & 8 $\times$ A100/H100 \\
Training epochs & 2 \\
Save frequency & every 20 steps \\
\midrule
\multicolumn{2}{l}{\textit{\modelname-specific}} \\
$\lambda_{\text{pos}}$ (cohesion) & 0.001 \\
$\lambda_{\text{sep}}$ / $\lambda_{\text{div}}$ / $\lambda_{\text{neg}}$ & 0 \\
Layer index & $-1$ (last layer) \\
Anchor position & pre\_boxed\_last\_token \\
Warmup start & step 0 \\
Warmup steps & 60 \\
\bottomrule
\end{tabular}
\end{table}

\paragraph{Evaluation Settings.}
For greedy evaluation (pass@1), we use temperature $= 0$ with a maximum generation length of 32,768 tokens and thinking mode disabled.
For pass@$k$ evaluation, we sample $n = 32$ responses per prompt with temperature $= 0.2$.
Pass@$k$ is computed using the unbiased estimator~\cite{chen2021evaluating}:
\[
\text{pass@}k = 1 - \frac{\binom{n-c}{k}}{\binom{n}{k}}
\]
where $n$ is the total number of samples and $c$ is the number of correct samples.
All evaluations use tensor parallelism of 1 (greedy) or 2 (pass@32) with vLLM, and GPU memory utilization of 0.9.
We report the best checkpoint for each configuration, selected by greedy accuracy on the validation set.

\section{Case Study}
\label{app:case_study}

Table~\ref{tab:case_study} illustrates the ``knowing but not doing'' phenomenon that \modelname addresses.
On this geometry problem (AIME 2024 I, Problem 11), the DAPO baseline \textit{can} find the correct answer ($104$) in some rollouts (11/32), but its most frequent incorrect output is $291$, a plausible intermediate value.
With \modelname, the model's correct-answer rollout rate increases dramatically (26/32), and the dominant greedy output shifts to the correct answer.
This demonstrates how alignment loss consolidates the ``correct decision'' representation, making the model's greedy behavior align with its best sampling capability.

\section{Hidden States vs.\ Logits Alignment}
\label{app:logits}

To verify that aligning hidden states is not equivalent to aligning output logits, we run an additional experiment replacing the last-layer hidden states with the pre-softmax logits in the alignment loss, keeping all other settings identical.
Table~\ref{tab:logits} shows that logit alignment performs substantially worse than hidden-state alignment and offers no improvement over the DAPO baseline.
This confirms that the alignment signal is specific to the hidden representation space and does not trivially reduce to output-level matching.

\begin{table}[h]
\centering
\small
\caption{Case study on AIME 2024 I Problem 11. \modelname shifts the dominant mode from an incorrect intermediate value to the correct answer.}
\label{tab:case_study}
\begin{tabular}{p{0.15\textwidth}p{0.55\textwidth}}
\toprule
\textbf{Problem} & Rectangles $ABCD$ and $EFGH$ are drawn such that $D,E,C,F$ are collinear. Also, $A,D,H,G$ all lie on a circle. If $BC=16$, $AB=107$, $FG=17$, and $EF=184$, what is the length of $CE$? \\
\midrule
\textbf{Ground truth} & $\boxed{104}$ \\
\midrule
\textbf{DAPO baseline} & 11/32 correct; most frequent wrong answer: $291$ (8/32) \\
\textbf{\modelname} & 26/32 correct; wrong answers scattered (no dominant mode) \\
\bottomrule
\end{tabular}
\end{table}

\begin{table}[h]
\centering
\small
\caption{Hidden-state vs.\ logit alignment (Qwen3-4B, anchor token, $\lambda=0.001$).}
\label{tab:logits}
\begin{tabular}{lcccc}
\toprule
\textbf{Method} & \textbf{AIME24} & \textbf{AIME26} & \textbf{AMC24} & \textbf{HMMT} \\
\midrule
DAPO (baseline) & 48.12 & 40.31 & 58.75 & 19.79 \\
Logit (pre-softmax) & 46.67 & 45.42 & 58.54 & 19.90 \\
Hidden-state (ours) & \textbf{59.17} & \textbf{50.31} & \textbf{62.15} & \textbf{27.08} \\
\bottomrule
\end{tabular}
\end{table}

\end{document}